\documentclass[10pt,twocolumn,letterpaper]{article}

\usepackage{cvpr}
\usepackage{times}
\usepackage{epsfig}
\usepackage{graphicx}
\usepackage{amsmath}
\usepackage{amssymb}
\usepackage{siunitx}
\usepackage{xfrac}

\usepackage[breaklinks=true,bookmarks=false]{hyperref}

\cvprfinalcopy 

\def\cvprPaperID{****} 

\ifcvprfinal\fi
\begin{document}

\title{Squareplus: A Softplus-Like Algebraic Rectifier}

\author{Jonathan T. Barron\\
{\tt\small barron@google.com}
}

\maketitle

\newcommand{\softplus}{\operatorname{softplus}}
\newcommand{\fn}{\operatorname{squareplus}}
\newcommand{\relu}{\operatorname{relu}}
\newcommand{\bspecial}{4\ln^2 2}

\begin{abstract}
We present squareplus, an activation function that resembles softplus, but which can be computed using only algebraic operations: addition, multiplication, and square-root. Because squareplus is $\sim\!6\times$ faster to evaluate than softplus on a CPU and does not require access to transcendental functions, it may have practical value in resource-limited deep learning applications.
\end{abstract}

\begin{figure}[t!]
    \begin{flushright}
    \includegraphics[width=3.235in]{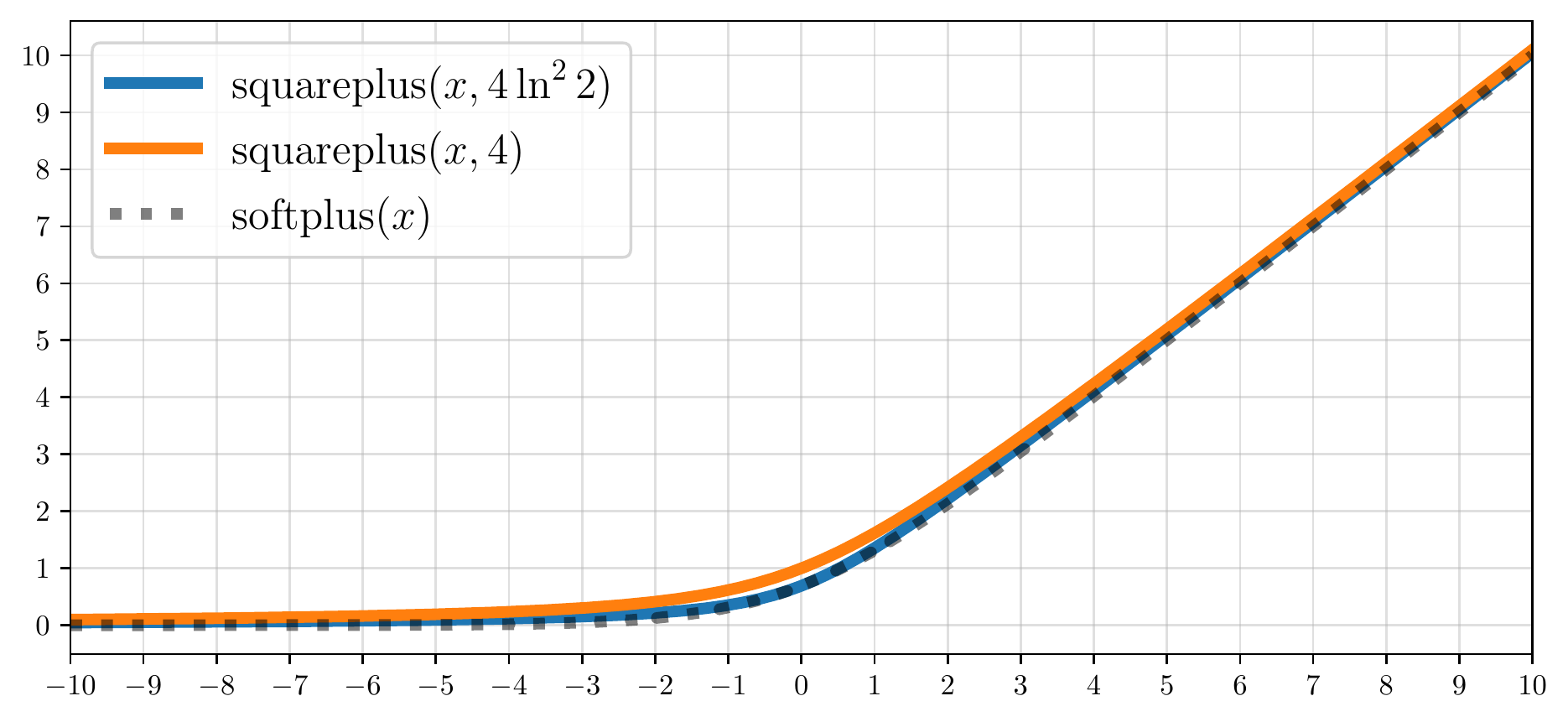}
    \includegraphics[width=\linewidth]{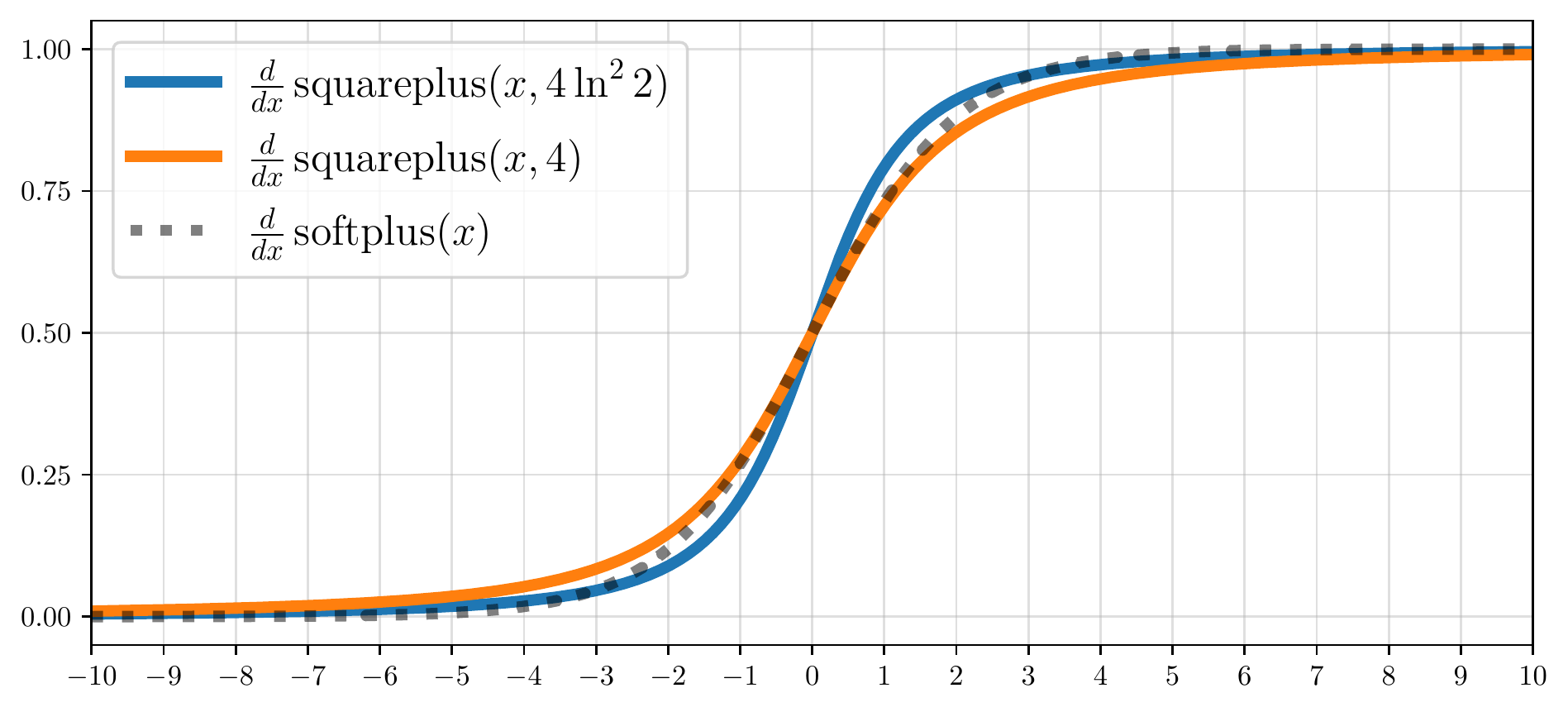}
    \includegraphics[width=\linewidth]{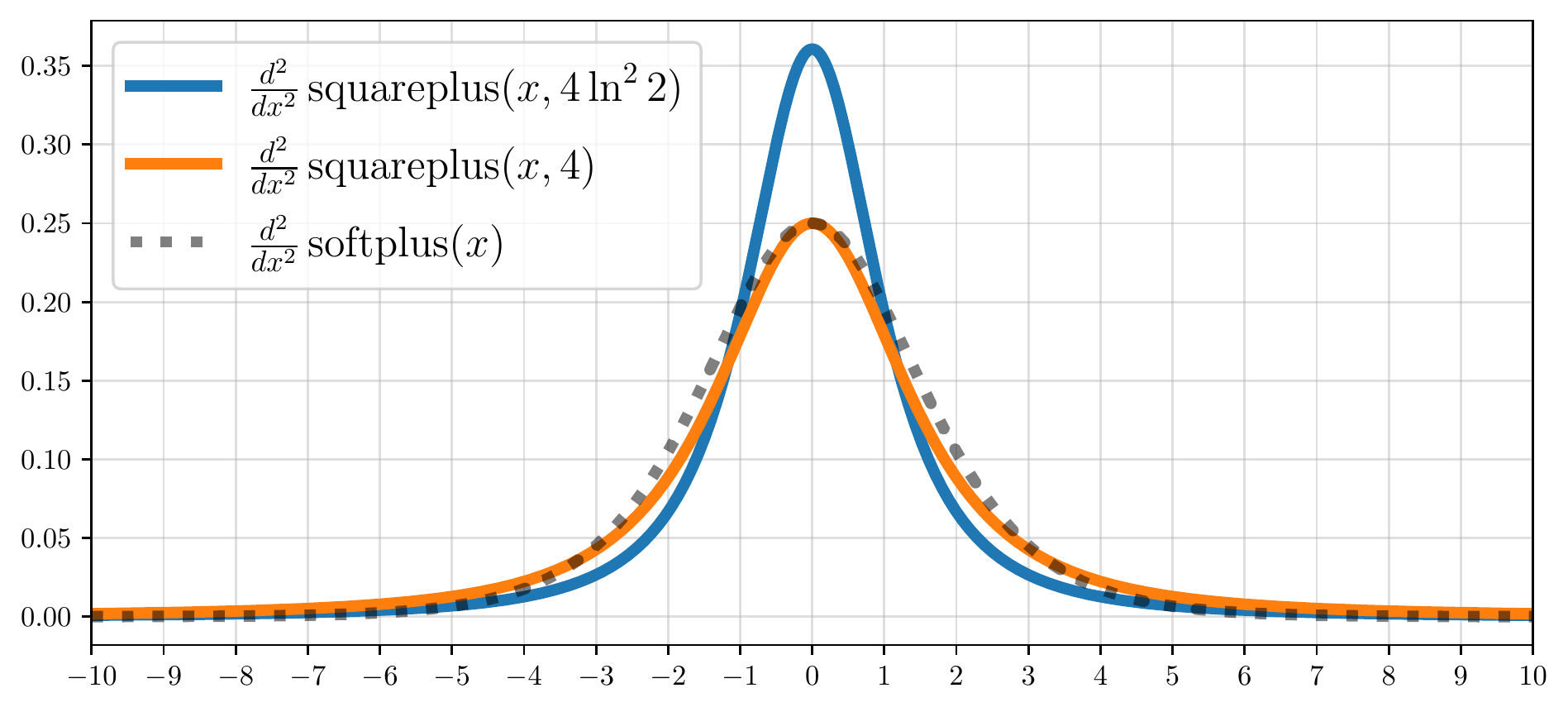}
    \end{flushright}
    \caption{
    A visualization of softplus and two instances of squareplus with different values of the $b$ hyperparameter, as well as their first and second derivatives. Squareplus approximates softplus when $b=\bspecial$, and approximates its second derivative when $b=4$.
    }
    \label{fig:plots}
\end{figure}

Activation functions are a central building block of deep learning architectures. The specific non-linearity applied at each layer of a neural network influences training dynamics and test-time accuracy, and is a critical tool when designing architectures whose outputs must lie within some range. When constraining a layer's output to be non-negative, a ubiquitous practice is to apply a ReLU activation~\cite{Fukushima1969VisualFE, glorot11a, Malik90}:
\begin{equation}
\relu(x) = \max(x, 0)
\end{equation}
Though ReLU ensures a non-negative output, it has two potential shortcomings: its gradient is zero when $x \leq 0$, and is discontinuous at $x=0$. If smooth or non-zero gradients are desired, a softplus~\cite{softplus} is often used in place of ReLU:
\begin{equation}
\softplus(x) = \log(\exp(x) + 1)
\end{equation}
Softplus is an upper bound on ReLU that approaches ReLU when $|x|$ is large but, unlike ReLU, is $C^\infty$ continuous. Though softplus is an effective tool, it too has some potential shortcomings: 1) it is non-trivial to compute efficiently, as it requires the evaluation of two transcendental functions, and 2) a naive implementation of softplus is numerically unstable when $x$ is large (a problem which can be straightforwardly ameliorated by returning $x$ as the output of $\softplus(x)$ when $x \gg 0$). Here we present an alternative to softplus that does not have these two shortcomings, which we dub ``squareplus'':
\begin{equation}
\fn(x, b) = \frac{1}{2} \left(x + \sqrt{x^2 + b}\right)
\end{equation}
Squareplus is defined with a hyperparameter $b \geq 0$ that determines the ``size'' of the curved region near $x=0$.
See Figure~\ref{fig:plots} for a visualization of squareplus (and its first and second derivatives) for different values of $b$, alongside softplus. Squareplus shares many properties with softplus: its output is non-negative, it is an upper bound on ReLU that approaches ReLU as $|x|$ grows, and it is $C^\infty$ continuous. However, squareplus can be computed using only algebraic operations, making it well-suited for settings where computational resources or instruction sets are limited. Additionally, squareplus requires no special consideration to ensure numerical stability when $x$ is large.

The first and second derivatives of squareplus are:
\begin{align}
\frac{d}{dx}\,\fn(x, b) &= \frac{1}{2}\left( 1 + \frac{x}{\sqrt{x^2 + b}} \right) \\
\frac{d^2}{dx^2}\,\fn(x, b) &= \frac{1}{2} \left( \frac{b}{(x^2 + b)^{\sfrac{3}{2}}} \right)
\end{align}
Like squareplus itself, these derivatives are algebraic and straightforward to compute efficiently. Analogously to how the derivative of a softplus is the classic logistic sigmoid function, the derivative of a squareplus is the ``algebraic sigmoid'' function  $x/\sqrt{x^2+1}$ (scaled and shifted accordingly). And analogously to how the second derivative of a softplus is the PDF of a logistic distribution, the second derivative of a squareplus (with $b=2$) is the PDF of Student's t-distribution (with $\nu=2$).

Specific values of the $b$ hyperparameter yield certain properties.
When $b=0$, squareplus reduces to ReLU:
\begin{equation}
\fn(x, 0) = \frac{x + |x|}{2}  = \relu(x)
\end{equation}
By setting $b=\bspecial$ we can approximate the shape of softplus near the origin:
\begin{equation}
\fn( 0, \bspecial) = \softplus(0)    
\end{equation}
This is also the lowest value of $b$ where squareplus's output is always guaranteed to be larger than softplus's output:
\begin{equation}
    \forall_{b \geq \bspecial} \,\,\, \fn(x, b) \geq \softplus(x)
\end{equation}
Setting $b=4$ causes squareplus's second derivative to approximate softplus's near the origin, and gives an output of $1$ at the origin (which the user may find intuitive):
\begin{gather}
\frac{d^2}{dx^2}\,\fn(0, 4) = \frac{d^2}{dx^2}\softplus(0) = \frac{1}{4} \\
\fn(0, 4) = 1
\end{gather}
For all valid values of $b$, the first derivative of squareplus is $\sfrac{1}{2}$ at the origin, just as in softplus:
\begin{equation}
\forall_{b \geq 0}  \,\,\ \frac{d}{dx} \fn(0, b) = \frac{d}{dx} \softplus(0) = \frac{1}{2}
\end{equation}
The $b$ hyperparameter can be thought of as a scale parameter, analogously to how the offset in Charbonnier/pseudo-Huber loss can be parameterized as a scale parameter~\cite{BarronCVPR2019, Charb}. As such, the same activation can be produced by scaling $x$ (and un-scaling the activation output) or by changing $b$:
\begin{equation}
    \forall_{a > 0} \,\,\, \frac{\fn(ax, b)}{a} = \fn\!\left(x, \frac{b}{a^2}\right)
\end{equation}

\begin{figure}[t]
    \centering
    \includegraphics[width=\linewidth]{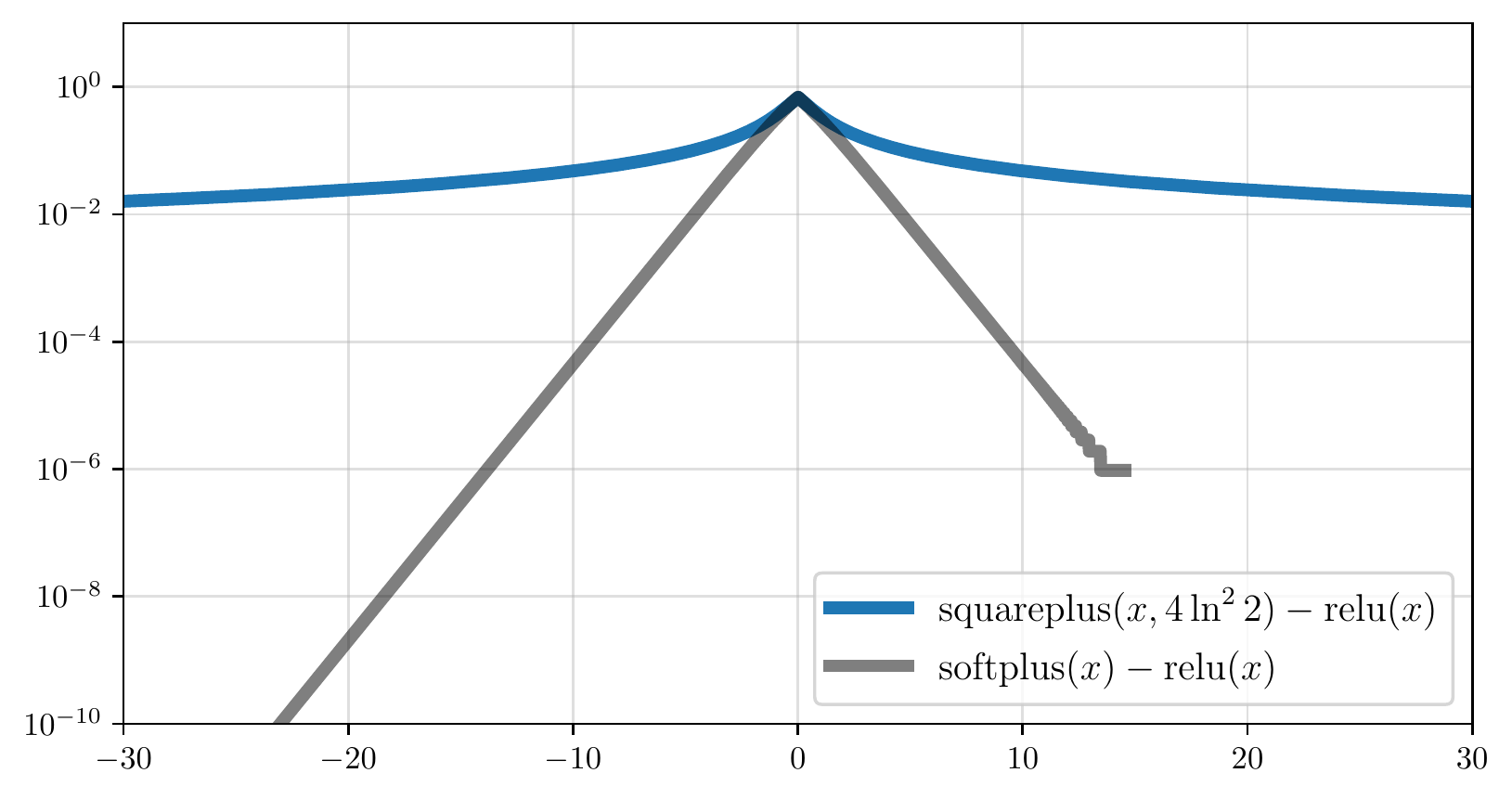}
    \caption{
    By visualizing the difference between squareplus/softplus and ReLU we see that squareplus approaches ReLU more slowly than softplus. The gray line terminates when softplus breaks down at $x \approx 15$, due to numerical instability.
    }
    \label{fig:log_plots}
\end{figure}

Though squareplus superficially resembles softplus, when $|x|$ grows large squareplus approaches ReLU at a significantly slower rate than softplus. This is visualized in Figure~\ref{fig:log_plots}, where we plot the difference between squareplus/softplus and ReLU. This figure also demonstrates the numerical instability of softplus on large inputs, which is why most softplus implementations return $x$ when $x \gg 0$. Similarly to this slow asymptotic behavior of the function itself, the gradient of squareplus approaches zero more slowly than that of softplus when $x \ll 0$. This property may be useful in practice, as "dying" gradients are often undesirable, but presumably this is task-dependent.

\begin{table}[b!]
\centering
\begin{tabular}{l|cc}
                         & CPU      & GPU \\ \hline
Softplus~\cite{softplus} (JAX impl.)     & \SI{3.777}{\milli\second} & \SI{1.120}{\milli\second}  \\
Softplus~\cite{softplus} (naive impl.)   & \SI{2.836}{\milli\second} & \SI{1.118}{\milli\second}  \\
ELU~\cite{Clevert2016FastAA}                      & \SI{2.040}{\milli\second} & \SI{1.120}{\milli\second}  \\
Swish/SiLU~\cite{ELFWING20183, GELU, Swish}          & \SI{1.234}{\milli\second} & \SI{1.113}{\milli\second}  \\
Relu~\cite{Fukushima1969VisualFE, glorot11a, Malik90}                     & \SI{0.598}{\milli\second} & \SI{1.069}{\milli\second} \\
Squareplus               & \SI{0.631}{\milli\second} & \SI{1.074}{\milli\second} 
\end{tabular}
\caption{
Runtimes on a CPU (for 1 million inputs) and a GPU (for 100 MM inputs) using JAX~\cite{jax2018github}. The ``naive implementation'' of softplus omits the special-casing necessary for softplus to produce finite values when $x$ is large.
}
\label{table:runtimes}
\end{table}

As shown in Table~\ref{table:runtimes}, on a CPU squareplus is $\sim\!6\times$ faster  than softplus, and is comparable to ReLU. On a GPU, squareplus is only $10\%$ faster than softplus, likely because all rectifiers are limited by memory bandwidth rather than computation in this setting. This suggests that squareplus may only be a desirable alternative to softplus in situations in which compute resources are limited, or when a softplus cannot be used --- perhaps because $\exp$ and $\log$ are not supported by the hardware platform.

\phantom{x}

\noindent {\bf Acknowledgements:} Thanks to the Twitter community for their helpful feedback to \url{https://twitter.com/jon_barron/status/1387167648669048833}

{\small
\bibliographystyle{ieee_fullname}
\bibliography{bib}

\begin{thebibliography}{10}\itemsep=-1pt

\bibitem{BarronCVPR2019}
Jonathan~T. Barron.
\newblock A general and adaptive robust loss function.
\newblock {\em CVPR}, 2019.

\bibitem{jax2018github}
James Bradbury, Roy Frostig, Peter Hawkins, Matthew~James Johnson, Chris Leary,
  Dougal Maclaurin, George Necula, Adam Paszke, Jake Vander{P}las, Skye
  Wanderman-{M}ilne, and Qiao Zhang.
\newblock {JAX}: composable transformations of {P}ython+{N}um{P}y programs,
  2018.

\bibitem{Charb}
P. Charbonnier, L. Blanc-Feraud, G. Aubert, and M. Barlaud.
\newblock Two deterministic half-quadratic regularization algorithms for
  computed imaging.
\newblock {\em ICIP}, 1994.

\bibitem{Clevert2016FastAA}
Djork-Arn{\'e} Clevert, Thomas Unterthiner, and S. Hochreiter.
\newblock Fast and accurate deep network learning by exponential linear units
  ({ELU}s).
\newblock {\em ICLR}, 2016.

\bibitem{softplus}
Charles Dugas, Yoshua Bengio, Fran\c{c}ois B\'{e}lisle, Claude Nadeau, and
  Ren\'{e} Garcia.
\newblock Incorporating second-order functional knowledge for better option
  pricing.
\newblock {\em NIPS}, 2000.

\bibitem{ELFWING20183}
Stefan Elfwing, Eiji Uchibe, and Kenji Doya.
\newblock Sigmoid-weighted linear units for neural network function
  approximation in reinforcement learning.
\newblock {\em Neural Networks}, 2018.
\newblock Special issue on deep reinforcement learning.

\bibitem{Fukushima1969VisualFE}
K. Fukushima.
\newblock Visual feature extraction by a multilayered network of analog
  threshold elements.
\newblock {\em IEEE Trans. Syst. Sci. Cybern.}, 1969.

\bibitem{glorot11a}
Xavier Glorot, Antoine Bordes, and Yoshua Bengio.
\newblock Deep sparse rectifier neural networks.
\newblock {\em NIPS}, 2011.

\bibitem{GELU}
Dan Hendrycks and Kevin Gimpel.
\newblock Bridging nonlinearities and stochastic regularizers with gaussian
  error linear units.
\newblock {\em CoRR}, abs/1606.08415, 2016.

\bibitem{Malik90}
Jitendra Malik and Pietro Perona.
\newblock Preattentive texture discrimination with early vision mechanisms.
\newblock {\em JOSA-A}, 1990.

\bibitem{Swish}
Prajit Ramachandran, Barret Zoph, and Quoc~V. Le.
\newblock Searching for activation functions.
\newblock {\em CoRR}, abs/1710.05941, 2017.

\end{thebibliography}
}

\end{document}